\documentclass[a4paper, 10pt, conference]{ieeeconf}      

\IEEEoverridecommandlockouts                              

\usepackage{array}
\usepackage{multirow} 
\usepackage{etoolbox}
\usepackage{subfig,graphicx}
\usepackage{utfsym}
\usepackage{bbding}
\usepackage{float}

\usepackage{adjustbox}
\usepackage{authblk}

\title{\LARGE \bf
Medical Visual Prompting (MVP): A Unified Framework for Versatile and High-Quality Medical Image Segmentation
}

\author[1]{Yulin Chen}
\author[2*]{Guoheng Huang\thanks{* Corresponding author.}}
\author[2]{Kai Huang\thanks{This work was supported in part by the Key Areas Research and Development Program of Guangzhou Grant 2023B01J0029, Science and technology research in key areas in Foshan under Grant 2020001006832, the Science and technology projects of Guangzhou under Grant 202007040006, the Guangdong Provincial Key Laboratory of Cyber-Physical System under Grant 2020B1212060069, the Guangdong Basic and Applied Basic Research Foundation under Grant 2023A1515012534, and the National Statistical Science Research Project of China (No. 2022LY096).}}
\author[2]{Zijin Lin}
\author[3]{Guo Zhong}
\author[4]{Shenghong Luo}
\author[5]{Jie Deng}
\author[67]{Jian Zhou}

\affil[1]{School of Automation, Guangdong University of Technology, Guangzhou, China}
\affil[2]{School of Computer Science and Technology, Guangdong University of Technology, Guangzhou, China}
\affil[3]{School of Information Science and Technology, Guangdong University of Foreign Studies, Guangzhou, China}
\affil[4]{Faculty of Science and Technology, University of Macau, Macau, China}
\affil[5]{Department of Otorhinolaryngology, The First Affiliated Hospital of Sun Yat-Sen University, Guangzhou, China}
\affil[6]{Guangdong Key Laboratory of Nasopharyngeal Carcinoma Diagnosis and Therapy,\protect\\ Sun Yat-sen University Cancer Center, Guangzhou, China.
}
\affil[7]{South China Hospital, Shenzhen University, Shenzhen, China}

\begin{document}

\maketitle
\thispagestyle{empty}
\pagestyle{empty}

\begin{abstract}

Accurate segmentation of lesion regions is crucial for clinical diagnosis and treatment across various diseases. While deep convolutional networks have achieved satisfactory results in medical image segmentation, they face challenges such as loss of lesion shape information due to continuous convolution and downsampling, as well as the high cost of manually labeling lesions with varying shapes and sizes. To address these issues, we propose a novel medical visual prompting (MVP) framework that leverages pre-training and prompting concepts from natural language processing (NLP). The framework utilizes three key components: Super-Pixel Guided Prompting (SPGP) for superpixelating the input image, Image Embedding Guided Prompting (IEGP) for freezing patch embedding and merging with superpixels to provide visual prompts, and Adaptive Attention Mechanism Guided Prompting (AAGP) for pinpointing prompt content and efficiently adapting all layers. By integrating SPGP, IEGP, and AAGP, the MVP enables the segmentation network to better learn shape prompting information and facilitates mutual learning across different tasks. Extensive experiments conducted on five datasets demonstrate superior performance of this method in various challenging medical image tasks, while simplifying single-task medical segmentation models. This novel framework offers improved performance with fewer parameters and holds significant potential for accurate segmentation of lesion regions in various medical tasks, making it clinically valuable.
\end{abstract}

\section{INTRODUCTION}
\begin{figure}[htb]
    \centering
    \includegraphics[width=\columnwidth]{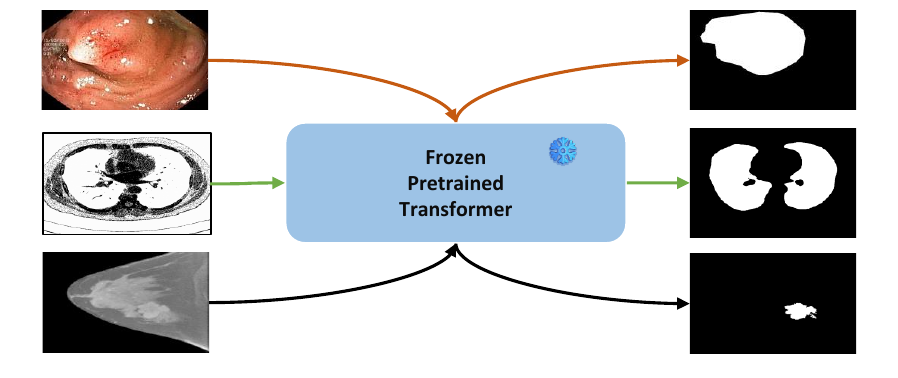}
    \caption{A novel framework for medical visual prompting, freezing the backbone, which can be applied to different medical data without updating the model.
}
\label{fig:wholemodel}
\end{figure}

Early detection of lesions plays a crucial role in clinical diagnosis. However, the unknown causes of most diseases pose challenges to accurate diagnosis. Advanced imaging techniques such as magnetic resonance imaging (MRI)~\cite{9878407,Chen2024-dg}, computed tomography (CT)~\cite{Gong2023GenerativeAF,Huang2023MRIS}, and endoscopy~\cite{10.1007/978-3-030-87193-2_12} aid physicians in making precise judgments. Nevertheless, medical image segmentation differs from ordinary RGB image segmentation~\cite{9412444} due to the unique physical properties of lesion regions and the variety of instruments used. Additionally, the high cost of manual annotation and the variations in lesion size and shape across different tasks further complicate the process. Visual prompts offer additional guidance, enhancing the model's understanding of image structures and features, facilitating the learning of generic features, and bolstering the model's generalization across various datasets~\cite{Wang2023FVPFV,10365931}. However, the complexity and diversity of medical imaging data, particularly in fine-grained segmentation, necessitate innovative approaches to effectively learn information from different datasets and guide segmentation networks to improve segmentation performance~\cite{Jia2022VisualPT}.

Traditional medical image segmentation methods often focus on specific tasks such as edge detection~\cite{9356353}. The advent of deep convolutional networks has greatly alleviated the limitations imposed by the diversity of medical images. Recent studies, such as MSNet and M2SNet~\cite{Zhao2021AutomaticPS,Zhao2023M2SNetMI}, employ specialized U-Net units to extract features that vary across image levels and are suitable for various medical tasks. However, successive convolutional layers and downsampling operations in mainstream segmentation networks may compromise shape information~\cite{Peng2023MShNetMF}, leading to the loss of important details and over-parameterization, thereby reducing computational efficiency.

Inspired by the latest prompting techniques~\cite{ Jia2022VisualPT}, we propose the Super Pixel Guided Prompting (SPGP) framework, which utilizes the shape information of lesion regions under different datasets to enhance the segmentation network's sensitivity to the shape of medical images. Additionally, we introduce the Image Embedding Guided Prompting (IEGP) framework for extracting shape-friendly prompts for medical image features. Furthermore, the Adaptive Attention Mechanism Guided Prompting (AAGP) framework employs a trainable attention mechanism to improve the accuracy of object boundaries.

Our approach combines SPGP, IEGP, and AAGP to enable the segmentation network to better learn shape prompting information, enhancing the model's generalization capability for downstream tasks. We propose a novel Medical Visual Prompting (MVP) framework that enables segmentation for different disease tasks through patch embedding and superpixel components as prompts. We achieve superior performance on various challenging medical image segmentation tasks crucial for clinical diagnosis.
In summary, our main contributions are as follows:

\begin{itemize}
\item We propose novel frameworks for medical image segmentation: Super Pixel Guided Prompting (SPGP) and Image Embedded Guided Prompting (IEGP), which utilize visual prompts to learn prior shape information and guide the segmentation network.
\item We design the Adaptive Attention Mechanism Guided Prompting (AAGP) framework to efficiently adapt to all layers by utilizing trainable attention mechanisms to localize the prompts more precisely.
\item We introduce the Medical Visual Prompting (MVP) framework, which fuses SPGP and IEGP as prompts and further utilizes AAGP for augmented prompting.The framework is applicable to various medical datasets without requiring modification of model.
\end{itemize}


\begin{figure*}[ht]
\vspace{-0.7cm}
        \centering
    \includegraphics[width=\linewidth]{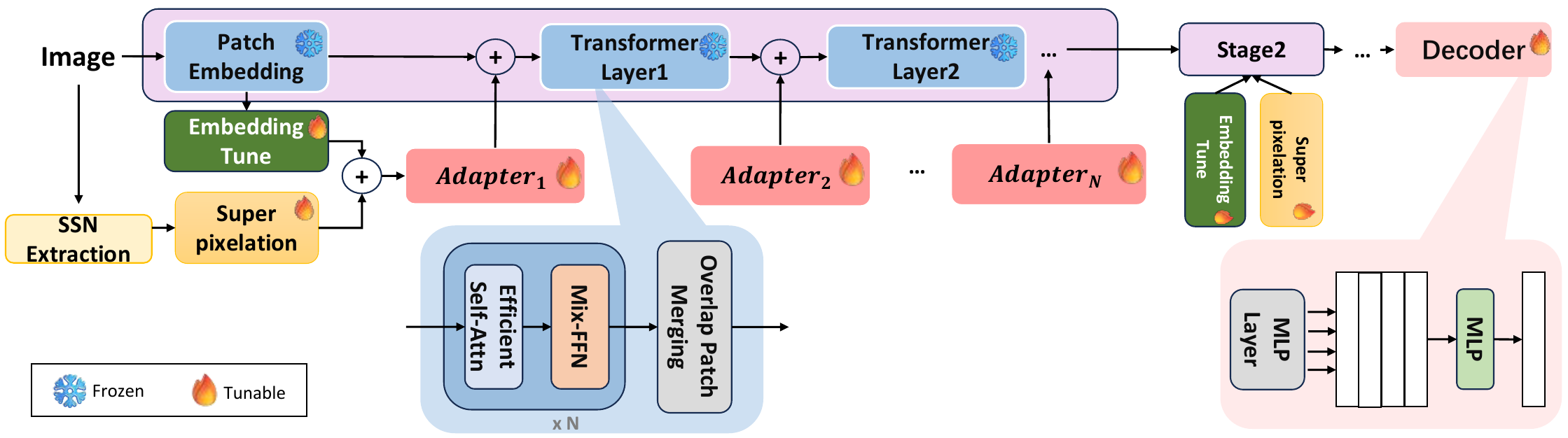}
    \caption{The architecture of the proposed effective medical visual prompting (EMVP). We use the Super-Pixel Guided Prompting (SPGP) and the Image Embedding Guided Prompting (IEGP) to tune the extracted features. The Adaptive Attention Mechanism Guided Prompting (AAGP) is designed to merge these features to focus on more effective visual prompting.}
    \label{fig:EMVPModel}
\end{figure*}

\section{RELATED WORK}

\subsection{Visual Prompting Tuning}
Natural Language Processing (NLP) has recently shifted from task-specific models to pre-trained model-based approaches with the rise of pre-trained language models (PLM)\cite{Rogers2020API}. Initially, pre-training and fine-tuning phases were used to adapt to downstream tasks\cite{Brown2020LanguageMA, Liu2021PretrainPA,liu2024depth}. To facilitate model prompting, a downstream task adaptation pre-training model was developed~\cite{Jia2022VisualPT, 9879120}. GPT-3~\cite{Brown2020LanguageMA} demonstrated high generalizability to downstream transfer learning tasks. The concept of prompting was extended to visual tasks with the introduction of Vision Transformer (ViT)~\cite{Dosovitskiy2020AnII,Luo2023DevignetHV,Chen2022ShadocnetLS,Li2023HighResolutionDS,Chen2023ALF}, which utilizes learnable embedding vectors adapted to new tasks. We employ a pre-trained Transformer model that introduces continuous embeddings after the input layer~\cite{9879120}, freezing the backbone network and updating only the prompt part during fine-tuning. Placing prompts before the transformer layer to guide the segmentation network improves model generalization to downstream tasks and efficiently prompts medical image segmentation~\cite{9879120}. While VPT focuses on recognition tasks~\cite{Jia2022VisualPT}, we concentrate on the best visual prompts for lesion regions under different datasets~\cite{8379359}.

\subsection{Medical Image Segmentation}
Medical image segmentation aims to highlight pathological structural changes in images. Traditional methods rely on machine learning~\cite{9356353}, but labeling is challenging. Fully Convolutional Network (FCN) extends classification to the pixel level~\cite{7478072}, while U-Net achieves edge feature extraction through feature splicing~\cite{Ronneberger2015UNetCN}, establishing the groundwork for future segmentation networks. UACANet focuses on uncertainty regions~\cite{Kim2021UACANetUA}. Recently, multiscale subtraction networks M2SNet~\cite{Zhao2023M2SNetMI} were proposed for various medical tasks. Despite the good performance of mainstream segmentation networks, successive convolutional layers and downsampling operations may compromise shape information~\cite{Peng2023MShNetMF}. In contrast, our work is inspired by the excellent performance of VPT~\cite{Jia2022VisualPT} in vision tasks. We fine-tune Transformer's input image patch embedding to adapt datasets from different tasks to a pre-trained large-scale model dataset as part of medical image prompts. Visual prompts contain abundant pixel information without modifying the parameters of the pre-trained model, guiding subsequent segmentation networks, and enhancing model generalization capability for medical datasets with limited data.

\section{METHOD}

\subsection{Overview of the MVP}
In this work, we propose an medical visual prompting (MVP) as shown in Fig~\ref{fig:EMVPModel}. Based on the hierarchical feature representation by SegFormer~\cite{Xie2021SegFormerSA}, it consists of the SPGP, IEGP, and AAGP together. Specifically, pre-trained visual transformer SegFormer on ImageNet is adapted to different medical image segmentation tasks. Throughout the training process, MVP keeps the backbone network frozen and trains only a small number of tunable parameters. The whole process is optimized by a Balanced Binary cross-entropy (BBCE) loss function.

\subsection{Super-Pixel Guided Prompting (SPGP)}\label{SPGP}
The core of SSN~\cite{10.1007/978-3-030-01234-2_22} is an end-to-end trainable superpixel sampling algorithm inspired by SLIC~\cite{6205760}. SSN consists of two parts: a deep network that generates pixel features, and a network that passes these features to a differentiable SLIC, which is a k-means clustering of image pixels in five-dimensional position and colour space (typically scaled $XY Lab$ space). First, it samples initial cluster (superpixel) centers ${{S}^0 \epsilon R^{m\times 5}}$ in five dimensions. Given these initial superpixel centers ${{S}^0}$, the two steps of pixel-superpixel association and superpixel center update are repeated $t$ times until convergence or a fixed number of iterations are performed. In this case, the pixel-superpixel association is to associate each pixel with the nearest superpixel center in the five-dimensional space, to determine to which new superpixel center each pixel $p$ belongs:
\begin{equation}
    H_{p}^{t}  = \mathop{\arg\min}\limits_{i\epsilon(0,...,m-1)}D(I_{p},S_{i}^{t-1}),\label{XX}
\end{equation}
where $D$ denotes the distance computation$D(a, b)= ||a-b||^{2}$, $m$ represents the number of pixels.

The superpixel center update calculates the average pixel feature $(XY+Lab)$ within each superpixel $i$ to obtain the new superpixel center $S^{t}$:
\begin{equation}
    S_{i}^{t}=\frac{1}{Z_{i}^{t}}\sum_{p|H_{p}^{t}=i}^{}I_{p},\label{yy}
\end{equation}
where $Z_{i}^{t}$ denotes the number of pixels in the superpixel cluster $i$.

Since pixel-superpixel association involves non-differentiable nearest neighbour calculation, it is removed from the hard distance $H\epsilon \{0,1,... ,m-1\}^{n\ast 1} $becomes soft distance $Q\epsilon R^{n\ast m}$, becomes non-differentiable to differentiable, and is updated to soft pixel-superpixel association $Q_{pi}^{t}$:
\begin{equation}
    Q_{pi}^{t}=e^{-D(I_{p},S_{i}^{t-1})}=e^{-||I_{p}-S_{i}^{t-1}||^2},\label{xy}
\end{equation}
and, superpixel center update:
\begin{equation}
    S_{i}^{t}=\frac{1}{Z_{i}^{t}}\sum_{p=1}^{n}Q_{pi}^{t}I_{p},\label{yx}
\end{equation}
where $Z_{i}^{t}=\sum_{p}^{}Q_{pi}^{t}$ is the normalization constant, for convenience, matrix the above formula:
\begin{equation}
    S^{t}=\hat{Q^t}^TI,\label{xxx}
\end{equation}
where $\hat{Q^t}$ is the normalized $Q^t$ of the column, shape $N\times M$ ($N$ is the number of pixels, $M$ is the number of superpixels).

The association matrix $Q$ obtained after the input image is processed by SSN, $\hat{q ^t}$ is the correlation feature of the superpixel part corresponding to each pixel. Input image is processed by SSN and then passed through the deep network again to obtain the intermediate feature $X$~\cite{9474911}. By each value of $X$ and the correlation matrix $\hat{Q^t}$, we can get the weight matrix $T$:
\begin{equation}
    T=\frac{X_{i}\hat{Q^t}}{\sum Q_{label=X-label}},\label{x1}
\end{equation}
where $Q_{label=X-label} $ is the sum of the associated features of the association matrix $\hat{ Q^t }$ belonging to the same superpixel block as $X_{i}$. 

Thus, the pixel-superpixel association matrix generated by SSN can be used to convert pixel-level features into super-pixelated feature components $X_{sp}$ :
\begin{equation}
    X_{sp} =\hat{ Q^t }XT,\label{yyy}
\end{equation}

\subsection{Image Embedding Guided Prompting (IEGP)}\label{IEGP}
The Image Embedding Guided Prompting (IEGP) module is designed to tune pre-trained patch embeddings. We refer to the EVP proposed by~\cite{Liu2023ExplicitVP,Chen2023MedPromptCP}, which projects a patch $X^{p}$ onto a space characterized by dimension $C_{seg} $. This shadow is then frozen and an adjustable linear layer $L_{pe}$ is added to project the original image embedding into the c-dimensional feature $X_{pe}\epsilon{R}^{c}$. Throughout, we introduce the scale factor $\gamma$ to control the tunable parameters of the patch embedding:
\begin{equation}
    X_{pe}=L_{pe}(X^{p}),with c =\frac{C_{seg}}{\gamma },\label{yyyy}
\end{equation}
where we introduce the scale factor $\gamma$ to control the tunable parameters.

\subsection{Adaptive Attention Mechanism Guided Prompting (AAGP)}\label{AAGP}
We propose the Adaptive Attention Mechanism Guided Prompting (AAGP) framework, which consists of Adapters that implement different functions to facilitate adaptive operations across all layers. For each $i$ adapter, we divide it into two parts to realize different functions. For the first part, we refer to the EVP proposed by ~\cite{Liu2023ExplicitVP}, taking $X_{pe}$ and $X_{sp}$ as inputs to the first part of the adapter and get the explicit visual prompts $P^{i}$ :
\begin{equation}
    P^{i}= MLP_{up}(GELU(MLP_{tune}^{i}(X_{pe}+X_{sp}))),\label{xyxy}
\end{equation}
where GELU is the GELU activation function. $MLP_{tune}^{i}$ is a linear layer used to generate hints for different tasks in each adapter. $MLP_{up}$ is the upper projection layer shared by all adapters to match the dimensions of the transformer features.

$P^{i}$ is the output prompt connected to each transformer layer.
Considering the specificity of different task lesion areas, the SPGP may be insensitive to the edge part and cannot be localized completely and accurately. We use the adaptive attention mechanism to focus on the edge information and optimize the prompt according to different tasks~\cite{Vaswani2017AttentionIA}, which improves the adaptability by prompting more accurately for the downstream tasks as the tasks keep changing. Thus, for the second part, we can obtain explicit visual prompts $P^{j}$ :
\begin{equation}
    P^{j}=Attention(x, H, W)=softmax(\frac{xH^{T}}{\sqrt{d_{h}}} )W,\label{yxyx}
\end{equation}
where $x$ is the original input image. $H$ and $W$ are the height and width of the input image respectively. $\sqrt{d_{h}}$is dimension of the processed image.

The final result of the adapter outputs the prompt $P^{k}$:
\begin{equation}
    P^{k}=P^{i}+P^{j},\label{xxxxyyyy}
\end{equation}



\section{EXPERIMENT}

\subsection{Dataset}
In this study, we evaluate our model on various datasets for three tasks: Endoscopic polyp segmentation, CT segmentation and MRI segmentation, three of them are private datasets and two are public datasets. Each sample includes an image and mask. A summary of the basic information of these datasets is illustrated in Table \ref{table1}.

\begin{table}[ht]
\centering
    \caption[Summary of datasets]{Summary of datasets considered in this work. We show the number of images in training (Train) and testing set (Test) for different datasets.}
    \label{table1}
\adjustbox{width=\columnwidth}{
\begin{tabular}{c|c|c|c|c}
\hline
Task                                                                                         & Dataset     & Train & Test & Resolution \\ \hline
\multirow{2}{*}{\begin{tabular}[c]{@{}c@{}}Endoscopic polyp \\ segmentation\end{tabular}} & Kvasir-SEG  & 700   & 300  & $620\times530$    \\
                                                                                             & Nasopharynx & 487   & 209  & $512\times512$    \\ \hline
\multirow{2}{*}{CT segmentation}                                                             & 2D-CT-Lung  & 187   & 80   & $512\times512$    \\
                                                                                             & ESOCT       & 4592  & 1730 & $512\times512$    \\ \hline
MRI segmentation                                                                             & BIMR        & 3659  & 421  & $385\times600$    \\ \hline
\end{tabular}
}
\end{table}

\textbf{Endoscopic polyp segmentation.} 
Kvasir-SEG is an open-access dataset of gastrointestinal polyp images and corresponding segmentation masks, manually labeled by a physician and then validated by an experienced gastroenterologist. Nasopharynx dataset approved by the Ethics Committee of the First Hospital of Sun Yat-sen University (No. [2023] 755) was collected by Dr. Jie Deng in 2021. Labels were annotated by two specialized endoscopists. In the experimental dataset, we normalized the resolution to 512$\times$512 pixels. We use the common used metrics such as $S_{m}$, MAE and $E_{\phi}$ to evaluate the performance.

\textbf{Computed tomography (CT) segmentation.} 
2D-CT-Lung is a 2D lung public dataset for medical image segmentation in Kaggle Data Science Bowl 2017. ESOCT is provided by Sun Yat-sen University Cancer
Center and is from 30 esophageal cancer patients.  CT scan of the chest is performed for each patientand then a total of 6322 CT slices are collected. The CT slices have 512$\times$512 pixels. Moreover, all the CT slices are labeled
manually by professional doctors. For evaluation metrics, we report Dice and Acc.

\textbf{Magnetic resonance imaging (MRI) segmentation.} 
BIMR is a privately owned breast dataset obtained by a novel 3D breast imaging device to provide important breast tissue information, provided by the Ethics Committee of Sun Yat-sen University Cancer Center (Approval No. B2019-016-01). Each sample in the BIMR dataset is in 3D high resolution and labeled by the corresponding medical experts. To obtain two-dimensional images, we sliced the three-dimensional samples, retaining the slices in which the lesion areas were obvious. We used the commonly used evaluation metrics: Dice and mIoU.

\subsection{Implementation Details}
All experiments were performed on a single NVIDIA GeForce GPU with 12G memory. All experiments were performed using SegFormer-B4~\cite{Xie2021SegFormerSA} pre-trained on the ImageNet-1k~\cite{5206848} dataset. At data loading, the images were selectively converted to grayscale images or RGB images based on the mask value. The AdamW optimizer was used for all experiments. For three different medical segmentation detection tasks, the learning rate was set to $5e^{-4}$. For five different datasets, the maximum epoch was set to 50, and the batch size of the training set was all 4.

\subsection{Evaluation metric}
The confusion matrix is a standardized format that expresses an evaluation of model accuracy. Thus, we can quantitatively assess the segmentation performance of the model by calculating evaluation metrics such as accuracy, dice coefficient (Dice) and mean joint intersection (mIoU). In addition, S-measure ($S_{m}$) and mean E-measure ($E_{\phi}$) are metrics for evaluating images based on their structural and contrast characteristics that can be applied to different scenarios. We additionally introduce Mean Absolute Error (MAE).

\subsection{Main Results}
\begin{figure*}[ht] 
\begin{minipage}[b]{0.3\linewidth}
    \centering
    \begin{table}[H]
    \centering
    \label{comparis}
    \caption{Quantitative results on Endoscopic segmentation.}
    \label{table2}
    \begin{tabular}{c|c|c}
    \hline
    \multirow{2}{*}{\begin{tabular}[c]{@{}c@{}}Method\end{tabular}} & {Kvasir-SEG} & {Nasopharynx} \\ \cline{2-3} 
                            & MAE↓   & MAE↓    \\ \hline
    U-Net~\cite{Ronneberger2015UNetCN}                   & 0.055  & 0.111    \\
    U-Net++~\cite{10.1007/978-3-030-00889-5_1}                 & 0.048  & 0.087     \\
    UACANet~\cite{Kim2021UACANetUA}               & 0.040    & \textbf{0.060}          \\
    M2SNet~\cite{Zhao2023M2SNetMI}                & 0.034   & 0.079          \\  
    FANet~\cite{9741842}                      & -   & 0.072           \\ \hline
    Ours                    & \textbf{0.032}  & 0.064      \\ \hline
    \end{tabular}
    \end{table}
\end{minipage}
\hfill
\begin{minipage}[b]{0.3\linewidth}
    \centering
    \begin{table}[H]
    \centering
    \caption{Quantitative results on CT segmentation.}
    \label{table3}
    \begin{tabular}{c|cc|cc}
    \hline
    \multirow{2}{*}{Method} & \multicolumn{2}{c|}{2D-CT-Lung} & \multicolumn{2}{c}{ESOCT} \\ \cline{2-5} 
                            & Dice↑          & Acc↑           & Dice↑        & Acc↑       \\ \hline
    U-Net~\cite{Ronneberger2015UNetCN}                   & 0.957          & 0.954          & 0.924        & 0.969           \\
    U-Net++~\cite{10.1007/978-3-030-00889-5_1}              & 0.968              & \textbf{0.970}          & 0.980             & 0.979             \\
    UACANet~\cite{Kim2021UACANetUA}      & 0.972             & 0.968          & 0.960             & 0.955   \\ 
    M2SNet~\cite{Zhao2023M2SNetMI}      & -             & 0.764          & 0.660             & \textbf{0.993}   \\  
    KiU-Net~\cite{9625988}               & \textbf{0.975}             & 0.945          & 0.977             & 0.982   \\ \hline
    Ours                    & 0.872          & 0.885            & \textbf{0.981}        & 0.986         \\ \hline
    \end{tabular}
    \end{table}
\end{minipage}
\hfill
\begin{minipage}[b]{0.3\linewidth}
    \centering
    \begin{table}[H]
    \centering
    \caption{Quantitative results on MRI segmentation.}
    \label{table4}
    \begin{tabular}{c|cc}
    \hline
    \multirow{2}{*}{Method} & \multicolumn{2}{c}{BIMR}                              \\ \cline{2-3} 
                        & \multicolumn{1}{l}{Dice↑} & \multicolumn{1}{l}{mIoU↑} \\ \hline
    UNet~\cite{Ronneberger2015UNetCN}                    & 0.762                     & 0.801                     \\
    UNet++~\cite{10.1007/978-3-030-00889-5_1}                  & 0.764                     & 0.803                     \\
    CMUNet~\cite{Tang2022CMUNeTAS}                  & 0.797                      & 0.826                    \\
    Deeplabv3~\cite{Chen2018EncoderDecoderWA}                                   & 0.850                        & 0.865                  \\
    FCN~\cite{7478072}                                 & 0.851                                & 0.866                 \\ \hline
    Ours                    & \textbf{0.975}                       & \textbf{0.981}                       \\ \hline
    \end{tabular}
    \end{table}
\end{minipage}

\end{figure*}

\begin{figure}[tb]
        \centering
    \includegraphics[width=0.5\textwidth]{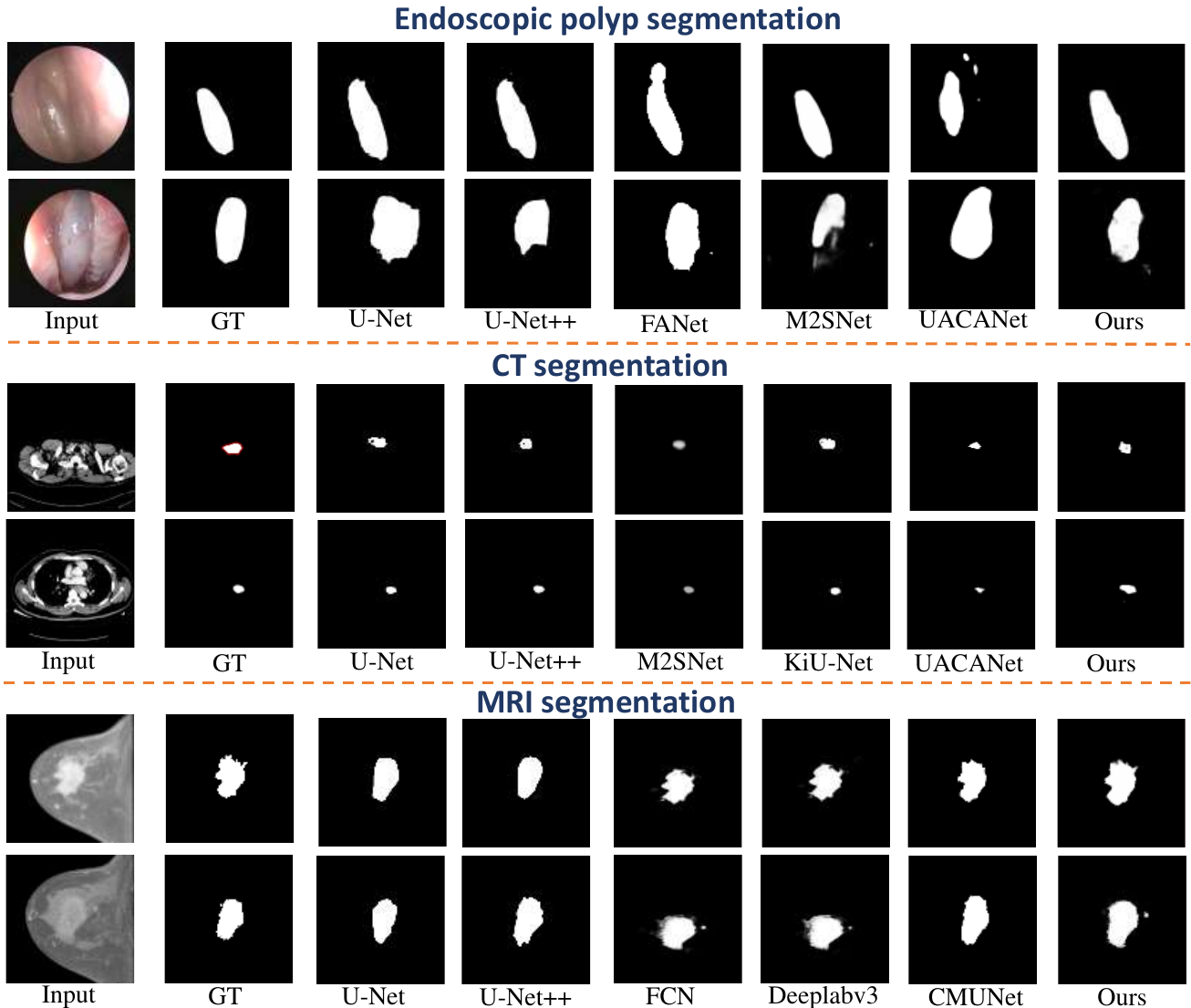}
    \caption{Comparison with other task-specific methods. We show the results for the Nasopharynx dataset (Top), the ESOCT dataset (Middle), the BIMR dataset (Bottom).}
    \label{fig:comparison}
\end{figure}


\begin{table*}[htb]
    \centering
    \caption{Ablation experiments were performed on the architectural design described in Figure xx. We evaluated it on three datasets for three different tasks. The proposed prompting strategy (Decoder + $X_{sp}$ + ($P_{i}$ + $P_{j}$) AAGP) is more effectively.}\label{table7}
\begin{tabular}{ccccc|c|ccc|c|cc}
\hline
\multicolumn{5}{c|}{\multirow{2}{*}{Models}}                                                                                                                                                                                                & \multirow{3}{*}{\begin{tabular}[c]{@{}c@{}}Trainable\\ Param.\\ (M)\end{tabular}} & \multicolumn{3}{c|}{\textbf{Polyp}} & \textbf{CT} & \multicolumn{2}{c}{\textbf{MRI}} \\
\multicolumn{5}{c|}{}                                                                                                                                                                                                                       &                                                                                   & \multicolumn{3}{c|}{Nasopharynx}    & ESOCT       & \multicolumn{2}{c}{BIMR}         \\ \cline{1-5}
\begin{tabular}[c]{@{}c@{}}Decoder\\ (no prompting)\end{tabular} & SPGP & \begin{tabular}[c]{@{}c@{}}AAGP\\$MLP_{tune}^{i}$\end{tabular} & \begin{tabular}[c]{@{}c@{}}AAGP\\ $MLP_{up}$\end{tabular} & \begin{tabular}[c]{@{}c@{}}AAGP\\ $P_{j}$\end{tabular} &                                                                                   & $S_{m}$↑        & MAE↓      & $E_{\phi}$↑      & Dice↑       & Dice↑           & mIoU↑          \\ \hline
\Checkmark                                                                &      &                                                        &                                                      &                                                   & 6.40                                                                              & 0.71       & 0.07      & 0.75       & -           & 0.52            & 0.29           \\
 &      & \Checkmark                                                      &                                                      &                                                   & 9.16                                                                              & 0.68       & 0.07      & 0.74       & 0.74        & 0.55            & 0.40           \\
 &      &                                                        & \Checkmark                                                    &                                                   & 10.20                                                                             & 0.69       & 0.07      & 0.75       & 0.74        & 0.52            & 0.37           \\
 &      & \Checkmark                                                      & \Checkmark                                                    & \Checkmark                                                 & 9.37                                                                              & 0.73       & 0.06      & 0.78       & 0.73        & 0.54            & 0.39           \\
 & \Checkmark    & \Checkmark                                                     & \Checkmark                                                    &                                                   & 9.37                                                                              & 0.55       & 0.13      & 0.62       & 0.73        & 0.11            & 0.06           \\
 & \Checkmark    & \Checkmark                                                     & \Checkmark                                                    & \Checkmark                                                 & 9.37                                                                              & \textbf{0.73}       & \textbf{0.06}      & \textbf{0.79}       & \textbf{0.87}        & \textbf{0.57}            & \textbf{0.42}           \\ \hline
\end{tabular}
\end{table*}
\textbf{Comparison with the task-specific methods.}
MVP performs well compared to task-specific methods. We report how our approach compares to other task-specific approaches in Table \ref{table2}, Table \ref{table3}, and Table \ref{table4}. Thanks to our more robust trunk and prompting strategies, MVP achieves comparable performance on 3 different tasks across the 5 datasets. However, compared to other well-designed domain-specific methods, MVP achieves commendable performance by introducing only a few tunable parameters on top of the frozen backbone. We also show a visual comparison with other methods on each task in Fig\ref{fig:comparison}. We can see that the proposed method predicts more accurate masks compared to other methods.

\subsection{Ablation Study}

We performed ablations to show the effectiveness of each component. Experiments were performed with a scaling factor of $r = 4$ except where noted.

\textbf{Architecture Design.}
To verify the effectiveness of the proposed visual prompting architecture, we modified it into different variants. As shown in Table \ref{table7}, the shared $MLP_{tune}^{i}$ tuning saves only a small number of parameters, but the performance drops significantly. Consistent performance gains are not obtained when using different $MLP_{up}$ in different adapters, in addition to introducing a large number of parameters. On the other hand, performance also drops dramatically when we remove $X_{sp}$ or $Adapter$, which indicates that they are both valid visual prompts.

\textbf{Tuning Stage.}
The aim of this study is to determine which stage contributes the most to the tuning of the prompt. To achieve this, we present a modified tuning approach where we vary the tunable stages in the SegFormer backbone. SegFormer has four stages for multi-scale feature extraction, and we mark the $Stage_{x}$ that adds tunable prompting as stage $x$. The results in Table \ref{table6} demonstrate that better performance can be achieved by adding tunable stages. Note that SegFormer-B4 utilizes 3, 8, 27 and 3 transformer blocks per stage. The effectiveness of MVP is positively correlated with the number of prompted transformer blocks.

\begin{table}[htb]
    \centering
    \setlength{\tabcolsep}{3pt}
    \caption{Ablation on the tuning stages in SegFormer. We conduct evaluations on three datasets for three different tasks. The performance of EMVP becomes better as the tuning stages increase.}\label{table6}
\begin{tabular}{c|c|ccc|c|cc}
\hline
\multirow{3}{*}{Method} & \multirow{3}{*}{\begin{tabular}[c]{@{}c@{}}Trainable\\Param.\\ (M)\end{tabular}} & \multicolumn{3}{c|}{\textbf{Polyp}} & \textbf{CT} & \multicolumn{2}{c}{\textbf{MRI}} \\
                        &                                                                                      & \multicolumn{3}{c|}{Nasopharynx}    & ESOCT       & \multicolumn{2}{c}{BIMR}         \\
                        &                                                                                      & $S_{m}$↑        & MAE↓       & $E_{\phi}$↑   & Dice↑       & Dice↑           & mIoU↑          \\ \hline
$Stage_{1}$            & 3.53                                                                                 & 0.68       & 0.07       & 0.72      & 0.74        & 0.54            & 0.38           \\
$Stage_{1, 2}$             & 4.02                                                                                 & 0.69       & 0.06       & 0.73      & 0.74        & 0.62            & 0.47           \\
$Stage_{1, 2, 3}$ & 5.93                                                                                 & 0.72       & 0.06       & 0.76      & 0.75        & \textbf{0.62}            & \textbf{0.47}           \\
$Stage_{1, 2, 3, 4}$  & 9.37                                                                                 & \textbf{0.73}       & \textbf{0.06}       & \textbf{0.79}      & \textbf{0.87}        & 0.57            & 0.42           \\ \hline
\end{tabular}
\end{table}

\section{CONCLUSIONS}
In this paper, we propose Medical Visual Prompting (MVP) to unify the solution for medical multitask segmentation. SPGP is the extraction of super-pixelated components from the original image. Purpose of IEGP is tuning pre-trained patch embedding. and fusing the two as hints. We also propose AAGP framework to make it efficiently adaptive in all layers. By using this approach, it enables the segmentation of lesion regions under different medical tasks and also, we find that the frozen vision converter backbone from ImageNet with a limited number of tunable parameters is also advanced in performance over other task-specific frozen vision converter backbones. The performance is also advanced compared to other task-specific methods, and this approach has proven to be effective and efficient in medical multitasking. In future work, we will extend our approach more deeply to problems related to the medical field in the hope that it will facilitate further exploration of visual prompts.








\bibliographystyle{IEEEtran}
\bibliography{ref}

\end{document}